\documentclass[twoside]{article}
\usepackage{papersty,palatino,epsfig,latexsym,natbib,amsmath,amssymb,amsfonts}

\begin{document}

\paperHeader{}{}{}{}{Novel crossover operators for subset selection}{A. Roy, J. D. Schaffer, and C. B. Laramee}
\title{\bf New crossover operators for multiple subset selection tasks}  

\author{\name{\bf Arnab Roy} \hfill \addr{aroy2@binghamton.edu}\\ 
        \addr{Department of Biomedical Engineering, Binghamton University, 
        Binghamton, 13902, U.S.A}
\AND
       \name{\bf J. David Schaffer} \hfill \addr{dschaffe@binghamton.edu}\\
        \addr{Department of Biomedical Engineering, Binghamton University, 
                Binghamton, 13902, U.S.A}
\AND
       \name{\bf Craig B. Laramee} \hfill \addr{claramee@binghamton.edu}\\
        \addr{Department of Biomedical Engineering, Binghamton University, 
                                Binghamton, 13902, U.S.A}
}

\maketitle
\begin{abstract}

We have introduced two crossover operators, MMX-BLX$^{exploit}$ and MMX-BLX$^{explore}$, for simultaneously solving multiple feature/subset selection problems where the features may have numeric attributes and the subset sizes are not predefined. These operators differ on the level of exploration and exploitation they perform; one is designed to produce convergence controlled mutation and the other exhibits a quasi-constant mutation rate. We illustrate the characteristic of these operators by evolving pattern detectors to distinguish alcoholics from controls using their visually evoked response potentials (VERPs). This task encapsulates two groups of subset selection problems; choosing a subset of EEG leads along with the lead-weights (features with attributes) and the other that defines the temporal pattern that characterizes the alcoholic VERPs. We observed better generalization performance from MMX-BLX$^{explore}$. Perhaps, MMX-BLX$^{exploit}$ was handicapped by not having a restart mechanism. These operators are novel and appears to hold promise for solving simultaneous feature selection problems.

\end{abstract}

\begin{keywords}
Subset selection,
Genetic algorithm,
Crossover operators,
Categorical genes,
Features with numeric attributes,
Evoked response potential,
Simultaneous feature selection
\end{keywords}

\section{Introduction}
The feature selection (FS) literature contains a wide variety of approaches designed to extract a subset of features that optimizes a given objective function \citep{Schaffer2005, Siedlecki1989, Kudo2000, Liu2005, Seok2004, Lucasius1992, Leardi1992, Oliveira2001, zhang2002, Yang1998, Mathias2000, bala1996using, radcliffe1992genetic, narendra1977branch}. These techniques are generally grouped into 2 approaches; the wrapper approach aims at choosing a feature subset that can improve the performance of a classifier, whereas, the filter approach consists of an objective function that exploits statistical properties of the data (e.g. information content, correlation coefficient) \citep{Liu2005}. Many empirical studies have reported that genetic algorithm (GA) based FS outperforms sequential search techniques for problems that involve a considerable number of features (typically $>$ 50) \citep{zhang2002, Seok2004}. Here, we present a novel crossover operator for the genetic algorithm (GA) based FS task and implement a wrapper model to illustrate its characteristics. Typically, bit-string chromosome representations have been used where the bits represent the absence/presence of a feature \citep{Yang1998, bala1996using, Leardi1992, Siedlecki1989, Seok2004},however, index representations (features encoded as numeric values) have also been implemented \citep{Lucasius1992, radcliffe1992genetic, Mathias2000}. Lucasius et al. \citep{Lucasius1992} introduced a crossover operator for index representation which was built upon the preservation of four basic properties of an encoded feature subset, identity, position, order, and adjacency, while transferring information from the parents to the daughter chromosome. Radcliffe \citep{radcliffe1992genetic} developed the RRR\footnote{Random Respectful Recombination} crossover operator for index representation and introduced the idea of respect, i.e. the child must inherit the common features between the parent chromosomes. Mathias et al. \citep{Mathias2000} have implemented similar encoding scheme and introduced the MMX\footnote{Mix and Match crossover}, the MMX-s\footnote{Mix and Match and sort crossover}, and the MSX\footnote{Mix and Swap crossover} crossover operators that maintain positive and negative respect in the absence of any mutation. The positive respect requires that the child must inherit common features between the parents and in order that the negative respect be maintained, the child should not contain features that are absent in both the parents. As a consequence of maintaining respect during the crossover, the above operators are known to produce convergence controlled variation (CCV) \citep{Mathias2000}; there is less variation among the chromosomes as the search converges. By introducing mutation to relax the requirement of negative respect, a convergence controlled mutation (CCM) can be achieved. For the purposes of this paper, we are interested to device two crossover operators, one that exhibits CCM and the other that has a quasi-constant mutation rate. Also, in order to facilitate encoding subsets of variable lengths, we decided to use index representation. \par Some engineering applications require that multiple FS problems be solved simultaneously. For example, Roy et al. \citep{Roy2013} developed a spike neural network (SNN) based classifier to characterize the alcoholic brain using visually evoked response potentials. This task involved simultaneously solving two FS problems; one that required choosing a correct subset of EEG leads along with the lead-weights (features with attributes) and the other that defined the temporal pattern to be detected. It can be inferred from the literature review that not much work has been done for simultaneously solving multiple FS problems, and where the features may contain numeric attributes. Here, we introduce two versions of MMX-BLX\footnote{Mix and match + blend crossover} operator, which is an extension of the MMX-SSS operator \citep{Schaffer2005}, to accommodate these requirements. The versions vary based on the level of exploration and exploitation they perform, hence we call them MMX-BLX$^{exploit}$ and MMX-BLX$^{explore}$. We illustrate the characteristic of the above crossover operators by evolving temporal pattern detectors to characterize the alcoholic brain.

\section{Crossover operators for numeric and categorical chromosomes}
Mathais et al. \citep{Mathias2000} introduced the concept of positive/negative respect and presented the MSX and the MMX crossover operators for subset selection problems where the subset size was predetermined. These operators were made more exploratory by compromising negative respect; the positions in the daughter chromosomes which consisted of unmatched elements, i.e. elements which are not common between parents, were mutated with a certain predefined probability. By introducing a numeric gene to encode the subset size (SSS), Schaffer et al. \citep{Schaffer2005} extended the MSX and the MMX crossover operators for FS problems where the SSS was not predefined but only an upper bound ($SSS_{max}$) on the SSS was given. In order to crossover the SSS gene, the BLX operator \citep{Eshelman92} was used. The SSS gene defined an acceptance boundary on the chromosome such that the encoded features on the right side of this boundary remained unexpressed; i.e. these features were not included in the subset. Therefore, in order to preserve the important features, MMX-SSS \citep{Schaffer2005} copied the common genes to offspring one position to the left of their parental positions. Here, we introduce the MMX-BLX$^{exploit}$ and the MMX-BLX$^{explore}$ crossover operators which are built upon the foundation of the MMX-SSS operator. We do not use a SSS gene; as a consequence, the order in which the features are encoded is irrelevant. Also, we allow the features to have multiple numeric attributes where the attributes of the common parental features are mated using the BLX operator \citep{Eshelman92}. Below we briefly present the BLX and the MMX-SSS crossover operators in order to introduce the important concepts necessary to explain the steps involved in MMX-BLX.


\begin{figure}[t]
\begin{center}
\centerline{ 
 \psfig{file=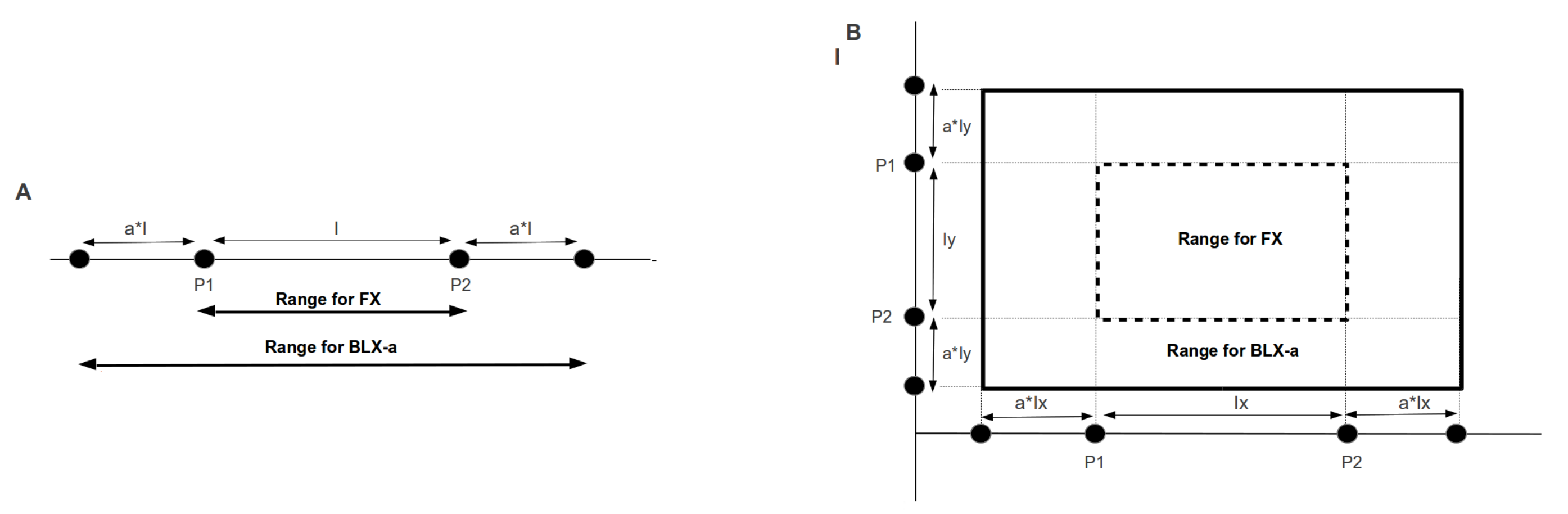,width=5.truein}
}
\end{center}
\caption{A: The BLX-$a$ operator generalizes FX operator by allowing the sampling range to expand beyond the parental allele values For $a$=0 BLX behaves similar to FX. B: The figure illustrate a 2-dimensional BLX and FX operation. FX samples within the area marked by the dotted rectangle and BLX samples within the area marked by the solid rectangle.}
\label{fig1}
\end{figure}

\subsection{BLX}
Radcliffe's flat crossover (FX) operator \citep{radcliffe1990} for numeric genes produces an offspring by uniformly picking a value between an interval defined by the parental allele-values (see Figure~\ref{fig1}a). The possible allele-values for the child are defined by a rectangular region for a 2-D problem,(see Figure~\ref{fig1}b), a region enclosed by a cuboid for a 3-D problem , and so on. Thus, the FX operator is respectful of this interval. Such a strategy is exploitative in nature and may lead to a premature convergence. Eshelman's and Schaffer's \citep{Eshelman92} BLX-$a$ ($\lbrace$ $a$  $\in$ $\mathbb{R}$ $\vert$ 0  $\le$ $a$ $\rbrace$) operator is a generalization of FX that allows the child to have allele-values in an extended region defined by the parameter $a$ (see Figure~\ref{fig1}a): BLX-$0$ is same as FX. If a child inherits an allele-value outside of the region bounded by the parent's allele-values, then it is considered to be a mutation-event. Clearly, the BLX-$a$ (for $a$ $>$ 0) operator is a crossover/mutation operator where the level of mutation is coupled with the degree to which the population is converged, as a result it exhibits CCM. Below, we have provided a pseudo-code for the BLX-$a$ operator.\\\\\\\\
\textbf{Pseudo-code for BLX-$a$ }\\\\
\textbf{Given:} 
\begin{enumerate}
\item[1.]{V1 = $(v_{1}^{1},v_{1}^{2},...,v_{1}^{\zeta})$ \tiny{$/\ast$ V1 is a $\zeta$-dimensional vector representing parent-1 allele values$\ast/$}}
\item[2.]{V2 = $(v_{2}^{1},v_{2}^{2},...,v_{2}^{\zeta})$ \tiny{$/\ast$ V2 is a $\zeta$-dimensional vector representing parent-2 allele values$\ast/$}}
\item[3.]{Vmax = $(v_{max}^{1},v_{max}^{2},...,v_{max}^{\zeta})$ \tiny{$/\ast$ A vector representing the maximum allowable allele values for the child$\ast/$}}
\item[4.]{Vmin = $(v_{min}^{1},v_{min}^{2},...,v_{min}^{\zeta})$ \tiny{$/\ast$ A vector representing the minimum allowable allele values for the child$\ast/$}}
\item[5.]{$a$, where $\lbrace$ $a$  $\in$ $\mathbb{R}$ $\vert$ 0  $\le$ $a$ $\rbrace$}
\end{enumerate}
\textbf{BLX(V1,V2,Vmax,Vmin,$a$)}
\begin{enumerate}
\item[1.]{op = () \tiny{$/\ast$ Define an empty output list op$\ast/$}}
\item[2.]{\textbf{For} counter: 1 to $\zeta$}
\item[3.]{\hspace{15pt}Range = $\|$ $v_{1}^{counter}$ - $v_{2}^{counter}$ $\|$ \tiny{$/\ast$ Calculate the interval $\ast/$}}
\item[4.]{\hspace{15pt}R1 = \textbf{Min}($v_{1}^{counter}$,$v_{2}^{counter}$) - Range$\ast$$a$ \tiny{$/\ast$ Define the lower bound $\ast/$}}
\item[5.]{\hspace{15pt}R2 = \textbf{Max}($v_{1}^{counter}$,$v_{2}^{counter}$) + Range$\ast$$a$  \tiny{$/\ast$ Define the upper bound $\ast/$}}
\item[6.]{\hspace{15pt}val = \textbf{Uniform-random}(R1,R2)  \tiny{$/\ast$ Generate uniform random number $\ast/$}}

\item[7.]{\hspace{15pt}\textbf{If }val $>$ $v_{max}^{counter}$ \textbf{then} val $=$ $v_{max}^{counter}$  \tiny{$/\ast$ Check for the upper bound $\ast/$}}

\item[8.]{\hspace{15pt}\textbf{If }val $<$ $v_{min}^{counter}$ \textbf{then} val $=$ $v_{min}^{counter}$  \tiny{$/\ast$ Check for the lower bound  $\ast/$}}
\item[9.]{\hspace{15pt}\textbf{Append}(op,val)  \tiny{$/\ast$ append val to the op list $\ast/$}}
\item[10.]{\textbf{Return}(op)}
\end{enumerate}
For the purposes of this paper the BLX operation for two scalar values, $w1$ and $w2$, with upper and lower bounds, $upper$ and $lower$, will be represented as BLX(($w1$),($w2$),($upper$),($lower$),$a$), where the parenthesis around the scalar values represents that they have been converted to a 1-dimensional vector.

\subsection{MMX-SSS}
In MMX-SSS all chromosomes are of fixed length ($SSS_{max}$), however, the number of genes that can be expressed may vary; this is defined using the SSS gene. The steps involved in the MMX-SSS crossover operation are as follows \citep{Schaffer2005}:
\begin{enumerate}
\item[1.]{\textbf{Incest prevention:} Avoid mating a pair of parent chromosomes which are too similar. Similarity is evaluated by comparing the expressed part of the chromosome that could be inherited by the offspring.}
\item[2.]{\textbf{Maintain positive respect:} Copy all the genes (features) that are common between the parents to the offspring. These genes will be copied one position to the left of their parental position. Child-1 and child-2 will receive the common genes from parent-1 and parent-2, respectively. }
\item[3.]{\textbf{Maintain negative respect:} The uncommon (uniques) genes between the parents are stored in a separate data-structure and sampled without replacement to fill the remaining positions in the offspring.}
\item[4.]{\textbf{Crossover SSS gene:} Perform BLX using the parent SSS genes to generate an integer value for the child SSS-gene.}
\item[5.]{\textbf{Mutation:} In order to allow for more exploration, the positions in the offspring occupied by the inherited unique genes will be replaced by an allele value randomly generated within an allowed range based on a predefined mutation rate. Care should be taken so that the same feature is not encoded twice.}
\end{enumerate}
As a consequence of maintaining positive respect (during crossover), over many generations more chromosomes will contain common features (genes). Also, as the inherited common elements in an offspring are never mutated, this will result in a CCM. Clearly, CCM is an important property that both BLX and MMX-SSS share.

\subsection{MMX-BLX}

The MMX-BLX crossover operator is designed for simultaneously solving $N$ FS problems where the features may have multiple numeric attributes and the subset sizes are not predefined. In order to accommodate these requirements we had to introduce a complex chromosome structure that consists of $N$ sub-chromosomes, where the sub-chromosome-$i$ ($\lbrace$ $i$  $\in$ $\mathbb{Z}$ $\vert$ 1  $\le$ $i$ $\le$ $N$ $\rbrace$) encodes a subset for the $i^{th}$ FS task. The MMX-BLX operation involves 3 basic tasks: defining the length of an offspring's $i^{th}$ sub-chromosome, modifying the attributes of the parental features\footnote{A copy of the parental features along with their attributes are stored in a separate data-structure. The feature-attributes are then modified in this data-structure before transferring them to the offspring.}, and defining the rules by which the offspring shall inherit features for the $i^{th}$ FS task. All the parameters that are necessary for these tasks are listed in Table~\ref{table1}. Below, we have explained the steps for accomplishing these tasks.

\begin{table}[h]
\begin{center}

\begin{tabular}{|c|p{4.5 cm}|l|}
 \hline
Parameter & Definition &  \multicolumn{1}{|c|}{Constraint}\\ \hline\hline
$i$ & \raggedright \small{Will be used for indexing the feature selection task} &  $\lbrace$ $i$  $\in$ $\mathbb{Z}$ $\vert$ 1  $\le$ $i$ $\le$ $N$ $\rbrace$  \\ \hline
 $\alpha$ & \raggedright \small{Will be used for defining the offspring's sub-chromosomal lengths.} &  $\lbrace$ $\alpha$  $\in$ $\mathbb{R}$ $\vert$ 0  $\le$ $\alpha$ $\rbrace$  \\ \hline
 $\beta$ & \raggedright \small{Will be used for modifying the attributes of the common parental features.} &  $\lbrace$ $\beta$  $\in$ $\mathbb{R}$ $\vert$ 0  $\le$ $\beta$ $\rbrace$   \\ \hline 
 $\delta$ & \raggedright \small{Will be used for the mutation operation.} &  $\lbrace$ $\delta$  $\in$ $\mathbb{R}$ $\vert$ 0  $\le$ $\delta$ $\le$ 1 $\rbrace$   \\ \hline
 $\gamma$ & \raggedright \small{Will be used for generating the attribute vector for the unique features.} &  $\lbrace$ $\gamma$  $\in$ $\mathbb{R}$ $\vert$ 0  $\le$ $\gamma$ $\rbrace$  \\ \hline
 $L_{i}^{att}$ & \raggedright \small{The length of the attribute vector of the features that are part of the $i^{th}$ FS task.}&   $\lbrace$ $L_{i}^{att}$  $\in$ $\mathbb{Z}$ $\vert$ 0  $\le$ $L_{i}^{att}$ $\rbrace$  \\ \hline
 $Val_{i}^{att:max}$ & \raggedright \small{A list defining the upper bound on the attribute values for features that are part of the $i^{th}$ FS task.} &    \\ \hline
 $Val_{i}^{att:min}$ & \raggedright \small{A list defining the lower bound on the attribute values for features that are part of the $i^{th}$ FS task.} & $Val_{i}^{attribute:max}$ $\geqq$ $Val_{i}^{attribute:min}$ \\ \hline
  $Val_{i}^{att:datatype}$ & \raggedright \small{A list defining the datatype of the individual elements of the attribute vector of features that are part of the $i^{th}$ FS task.} &  Datatype can be integer or real\\ \hline
 $Val_{i}^{feature:max}$ & \raggedright \small{The max feature-index for the $i^{th}$ FS task.} &    \\ \hline
 $Val_{i}^{feature:min}$ & \raggedright \small{The min feature-index for the $i^{th}$ FS task.} &  $Val_{i}^{feature:max}$ $\ge$ $Val_{i}^{feature:min}$ \\ \hline
$L_{i}^{max}$ & \raggedright \small{The maximum subset size for the $i^{th}$ FS task.} &    \\ \hline
 $L_{i}^{min}$ & \raggedright \small{The minimum subset size for the $i^{th}$ FS task.} & \tiny{($Val_{i}^{feature:max}$ - $Val_{i}^{feature:min}$ + 1) $>$ $L_{i}^{max}$ $\ge$ $L_{i}^{min}$  $>$ 0}\\ \hline   
 $N$ & \raggedright \small{Total number of FS tasks} & $\lbrace$ $N$  $\in$ $\mathbb{Z}$ $\vert$ 0  $<$ $N$ $\rbrace$  \\ \hline
 $P1$ & \raggedright \small{Parent-1 chromosome.} &   \\ \hline
$P2$ & \raggedright \small{Parent-2 chromosome.} &    \\ \hline
 \hline 
\end{tabular}
\caption{\label{table1} The above table illustrates the parameters required by the MMX-BLX crossover operator. }
\end{center}
\end{table}

\subsubsection{Defining the length of an offspring}
The length of the sub-chromosome-$i$ of chromosome-$x$ ($L_{i}^{x}$) is defined by the number of features (genes) it encodes and may vary among chromosomes. The length of offspring's sub-chromosome-$i$ ($L_{i}^{child}$) is defined by performing BLX operation over a range bounded by the lengths of the parental sub-chromosome-$i$ (see Figure~\ref{fig1}a). This operation is formally defined as follows:

\begin{equation}
\upsilon_{i}^{min} = Min(L_{i}^{parent_{1}} - \alpha, L_{i}^{parent_{2}} - \alpha)
\end{equation}

\begin{equation}
\upsilon_{i}^{max} = Max(L_{i}^{parent_{1}} + \alpha, L_{i}^{parent_{2}} + \alpha)
\end{equation}

\begin{equation}
L_{i}^{child} = Round(BLX((\upsilon_{i}^{min}),(\upsilon_{i}^{max}),(L_{i}^{max}),(L_{i}^{min}),0))
\end{equation}
The Round function guarantees that the output of this function will be an integer.

\subsubsection{Modifying the attributes of the parental features}
In MMX-BLX, all features that are part of the $i^{th}$ FS task must have attribute vectors of same length: $L_{i}^{att}$. For a pair of parental chromosomes, for the $i^{th}$ FS task, the common, the unique, and the absent features (along with their attributes) are stored in 3 bags: bag of common (BC-$i$), bag of uniques (BU-$i$), and bag of absent (BA-$i$), and their attributes are modified\footnote{Attributes of the features stores in the bags are modified. The parental chromosomes remain unchanged.}. The bags will contain only 1 instance of a feature; there will be no duplicate copies. The process by which the attributes are modified are discussed below:


\begin{figure}[t]
\begin{center}
\centerline{ 
 \psfig{file=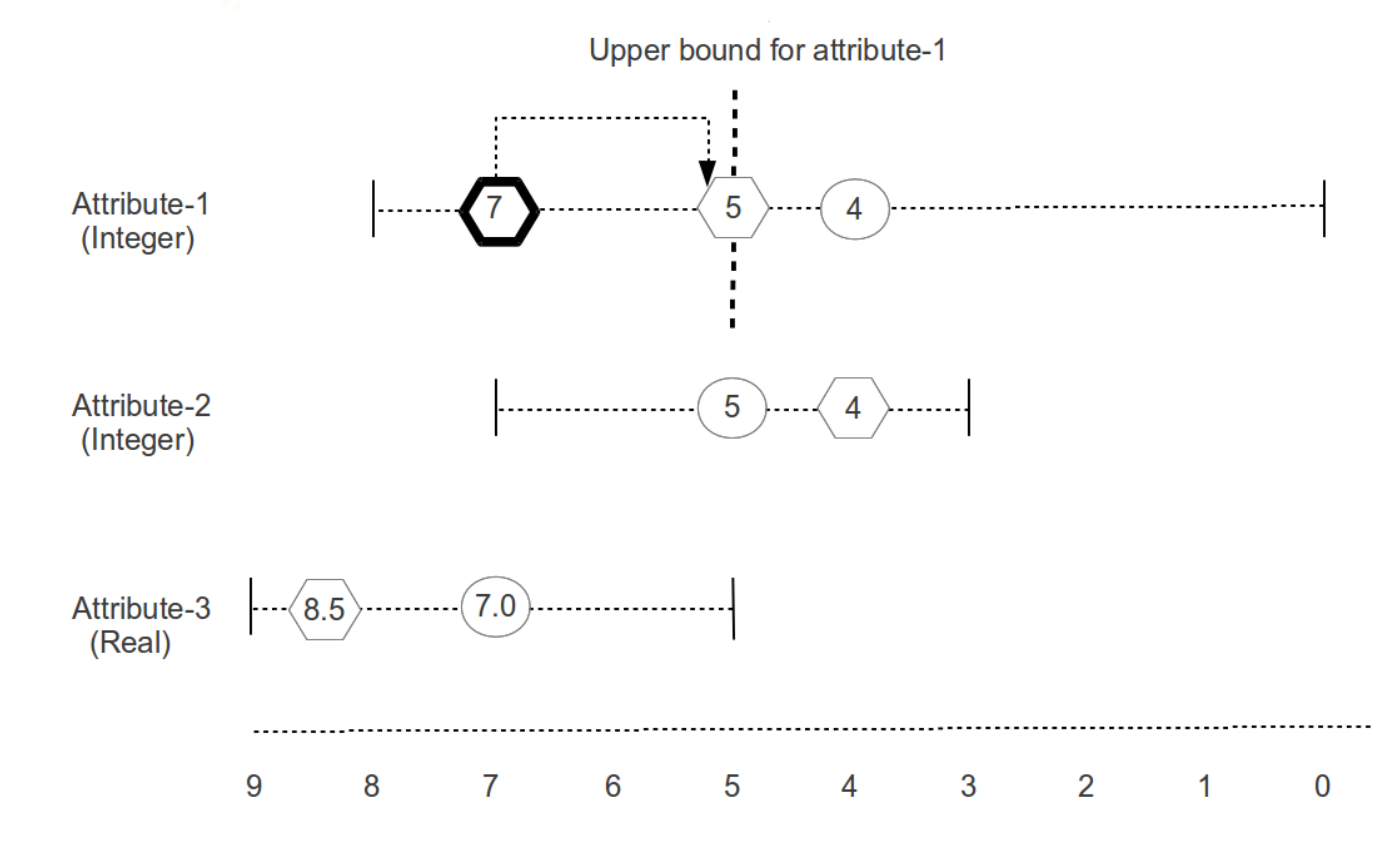,width=3.0truein}
}
\end{center}
\caption{The above figure illustrates the process by which new attribute vector (the hexagonal boxes) for the child is generated from the parental attribute-vector (the circles) of an unique feature. In the above figure the attribute-vector consists of 2 integer components and 1 real component. The dotted lines indicate the range within which the child attribute values will be generated. The range varies for different elements depending on the values of the corresponding elements of $Val_{i}^{att:max}$ and $Val_{i}^{att:min}$ vector, and the parameter $\gamma$. If any element of the child attribute-vector (the highlighted hexagonal box) exceeds the maximum/minimum allowable value (the vertical dotted lines), it will be updated to the threshold value it violated.}
\label{fig2}
\end{figure}

\begin{enumerate}
\item[1.]{For a common parental feature, $fcom_{i}$, the attribute-vectors, $Vcom_{fcom_{i}}^{parent_{1}}$ and $Vcom_{fcom_{i}}^{parent_{2}}$, are used for defining a region for the BLX operation to generate a new attribute-vector for the offspring (see Figure~\ref{fig1}b). This operation is formally defined as:
\begin{equation}
Vcom_{fcom_{i}}^{child} = BLX(Vcom_{fcom_{i}}^{parent_{1}},Vcom_{fcom_{i}}^{parent_{2}},Val_{i}^{att:max},Val_{i}^{att:min},\beta) 
\end{equation}
where, $fcom_{i}$ represents an instance of all common features for the $i{th}$ FS task.\\
}
\item[2.]{
 For an unique feature, $funq_{i}$, the attribute-vector, $Vunq_{funq_{i}}^{parent}$, is modified by performing BLX over a range (controlled by the parameter $\gamma$) that is defined by two vectors which are derived by incrementing and decrementing each element of $Vunq_{funq_{i}}^{parent}$ by an equal amount (see Figure~\ref{fig2}). The steps involved in this operation are shown below: 

\begin{equation}
V^{Range} = Val_{i}^{att:max} - Val_{i}^{att:min}
\end{equation}

\begin{equation}
V^{Lower} = Vunq_{funq_{i}}^{parent} - \frac{V^{Range}*\gamma}{2}
\end{equation}

\begin{equation}
V^{Upper} = Vunq_{funq_{i}}^{parent} + \frac{V^{Range}*\gamma}{2}
\end{equation}

\begin{equation}
Vunq_{funq_{i}}^{child} = BLX(V^{Upper},V^{Lower},Val_{i}^{att:max},Val_{i}^{att:min},0)
\end{equation}
where, $funq_{i}$ represents an instance of all unique features for the $i{th}$ FS task.\\
}
\item[3.]{ For an absent feature (features that are not present in either parents), $fabs_{i}$, the attribute-vector is set by performing BLX over a region that is bounded by 2 user-defined vectors (see Figure~\ref{fig1}b): $Val_{i}^{att:max}$ and $Val_{i}^{att:min}$. This operation is formally defined as:
\begin{equation}
 Vabs_{fabs_{i}}^{child} = BLX(Val_{i}^{att:max},Val_{i}^{att:min},Val_{i}^{att:max},Val_{i}^{att:min},0) 
\end{equation}
where, $fabs_{i}$ represents an instance of all absent features for the $i{th}$ FS task.
}
\end{enumerate}

\subsubsection{Inheriting features}
Below, we have presented the steps involved in the MMX-BLX$^{exploit}$ and the MMX-BLX$^{explore}$ crossover operation for the $i^{th}$ sub-chromosome. MMX-BLX operation produces a single offspring for a given pair of parents; in order to produce 2 offspring for a given parental pair, 2 independent MMX-BLX operations should be performed \footnote{In our experiments 2 offspring are produced per parental pair}. To generate an offspring, the following steps should be repeated for all sub-chromosomes.\linebreak\linebreak
\textbf{MMX-BLX$^{exploit}$}
\begin{enumerate}
\item[1.]{Determine the length ($L_{i}^{child}$) of the offspring's sub-chromosome-$i$ using the process discussed above.}
\item[2.]{If BC-$i$ is not empty then choose a feature (along with the attributes) from it randomly without replacement and place it in the offspring's sub-chromosome-$i$ else goto step-4.} 
\item[3.]{If the current length of the offspring's sub-chromosome-$i$ is less than $L_{i}^{child}$ then goto step-2.}
\item[4.]{Choose a feature (along with the attributes) randomly without replacement from BU-$i$ or BA-$i$ with a predefined probability $\delta$ and place it in the offspring's sub-chromosome-$i$. The parameter $\delta$ will become irrelevant when either BU-$i$ or BA-$i$ becomes empty.}
\item[5.]{If the current length of the offspring's sub-chromosome-$i$ is less than $L_{i}^{child}$ then goto step-4 else stop.}
\end{enumerate}
\textbf{MMX-BLX$^{explore}$}
\begin{enumerate}
\item[1.]{Determine the length ($L_{i}^{child}$) of the offspring's sub-chromosome-$i$ using the process discussed above.}
\item[2.]{If BC-$i$ and BA-$i$ are not empty then choose a feature (along with the attributes) randomly without replacement from BC-$i$ or BA-$i$ with a predefined probability $\delta$ and place it in the offspring's sub-chromosome-$i$. Then goto step-6.} 
\item[3.]{If BC-$i$ is empty and BA-$i$ is not, then choose a feature (along with the attributes) randomly without replacement from BU-$i$ or BA-$i$ with a predefined probability $\delta$ and place it in the offspring's sub-chromosome-$i$.  Then goto step-6.} 
\item[4.]{If BA-$i$ is empty and BC-$i$ is not, then choose a feature (along with the attributes) randomly without replacement from BC-$i$ and place it in the offspring's sub-chromosome-$i$.  Then goto step-6.} 
\item[5.]{If both BC-$i$ and BA-$i$ are empty then choose a feature (along with the attributes) randomly without replacement from BU-$i$ and place it in the offspring's sub-chromosome-$i$.  Then goto step-6.} 
\item[6.]{If the current length of the offspring's sub-chromosome-$i$ is less than $L_{i}^{child}$ then goto step-2 else stop.}
\end{enumerate}

\subsubsection{Summarizing MMX-BLX}
The MMX-BLX$^{exploit}$ operator places a greater emphasis on positive respect but allows for mutation to occur by sacrificing negative respect. By setting the parameter $\delta$ = 0 negative respect can be maintained. At the other extreme, by setting  $\delta$ = 1 the search can be made to be more exploratory. Regardless of the $\delta$ parameter value, MMX-BLX$^{exploit}$ always prioritizes positive respect. This aspect of MMX-BLX$^{exploit}$  is very similar to MMX-SSS. MMX-BLX$^{explore}$ however, sacrifices respect (both +ve and -ve) in order to allow for more exploration to occur. Clearly, MMX-BLX$^{exploit}$ should exhibit CCM, whereas MMX-BLX$^{explore}$ should have a quasi-constant mutation rate. MMX-BLX$^{explore}$ may be useful for problems that require more exploration or where the fitness function gradually changes over time. We believe that both operators can be made more effective if some form of incest prevention scheme can be applied. Here, we have not implemented such a scheme as defining a similarity measure for chromosomes that encode multiple feature subsets is not trivial and requires a separate investigation. In comparison to MMX-SSS where all chromosomes are of fixed length and the SSS gene defines the expressible part of the chromosome, here, chromosomes vary in size and there are no passive genes. Hence, the memory is more efficiently used. \par While simultaneously solving $N$ FS tasks, there may be an additional constraint that some of the sub-chromosomes must be of same length; i.e. the encoded subset sizes must be similar. In order to satisfy this constraint, we generate the lengths of the offspring's sub-chromosomes that encode these tasks using a parental sub-chromosome-$i$ that encodes one of these tasks. This is possible because all the parental sub-chromosomes encoding these tasks are of same lengths.

\section{The alcoholic classification task}

The screening procedures for alcoholism currently used in the clinics are mainly questionnaire based tests that involve queries related to social/family problems the subject may have faced due to drinking, guilt associated with the addiction, and pattern of alcohol consumption  \citep{Cherpitel1997, Kahan1996}. These tests are not very accurate and their results may vary with gender and race \citep{Cherpitel1997}. Tests that are based on detecting the physiological changes associated with the disease may not only be more accurate, but may also be more informative for clinical purposes. Recent MRI, fMRI, and EEG studies have reported occurrences of structural and functional changes in the alcoholic brain \citep{harper1998, harper1987, harper1985, george2004fro, Mann2001}. An EEG based alcoholic screening procedure may be clinically useful because of its portability, afford-ability , and good temporal resolution. Here, we will evolve a temporal pattern detector (TPD) that can characterize the alcoholic brain using its visually evoked response potentials (VERPs) which are recorded using electroencephalogram (EEG). The primary goal of the evolutionary process will be to select a set of EEG leads along with weights and to evolve the design specifications for the TPD. Below, we have explained the VERP dataset, the TPD technology, and the steps involved in evolving the TPDs for the alcoholic classification task.

\subsection{The VERP dataset}
The dataset consisted of VERPs from two groups of subjects: alcoholics and controls. Each subject was exposed to a visual stimulus (a picture), thus providing 1 second of signal (256 samples) from 62 EEG leads starting at the exposure to the picture. For a given subject, multiple such trials were performed; the number of trials per subject varied from 7 to 60. For a given trial, if any of the leads generated a signal that exceeded 100$\mu$V, it was discarded for containing blink artifacts. Subjects with fewer than 40 trials were not included in this study; 47 alcoholics and 31 controls were eligible. For each subject, 36 randomly chosen trials were used to calculate the average signal across each lead \footnote{62 average signals (representing 62 leads) each composed of 256 data points.} for training the classifier and another set of 36 randomly chosen trials were used for developing an average signal for the testing purposes; clearly, there will be many instances of overlapping trails. In order to reduce the extent of overlap, we could have used fewer trials for averaging, however, the signals would have been too noisy for the classification task. VERPs are much weaker than the background EEG activity and typically at-least 100 trails \footnote{with no artifacts} are required to generate an averaged signal with a respectable signal to noise ratio. Hence, we expect our training files to be noisy and therefore, we believe it is an interesting challenge for the evolution to find a robust temporal pattern with an appropriate tolerance level that can characterize the alcoholic VERPs.
 

\subsection{The temporal pattern detector}

\begin{figure}[t]
\begin{center}
\centerline{ 
 \psfig{file=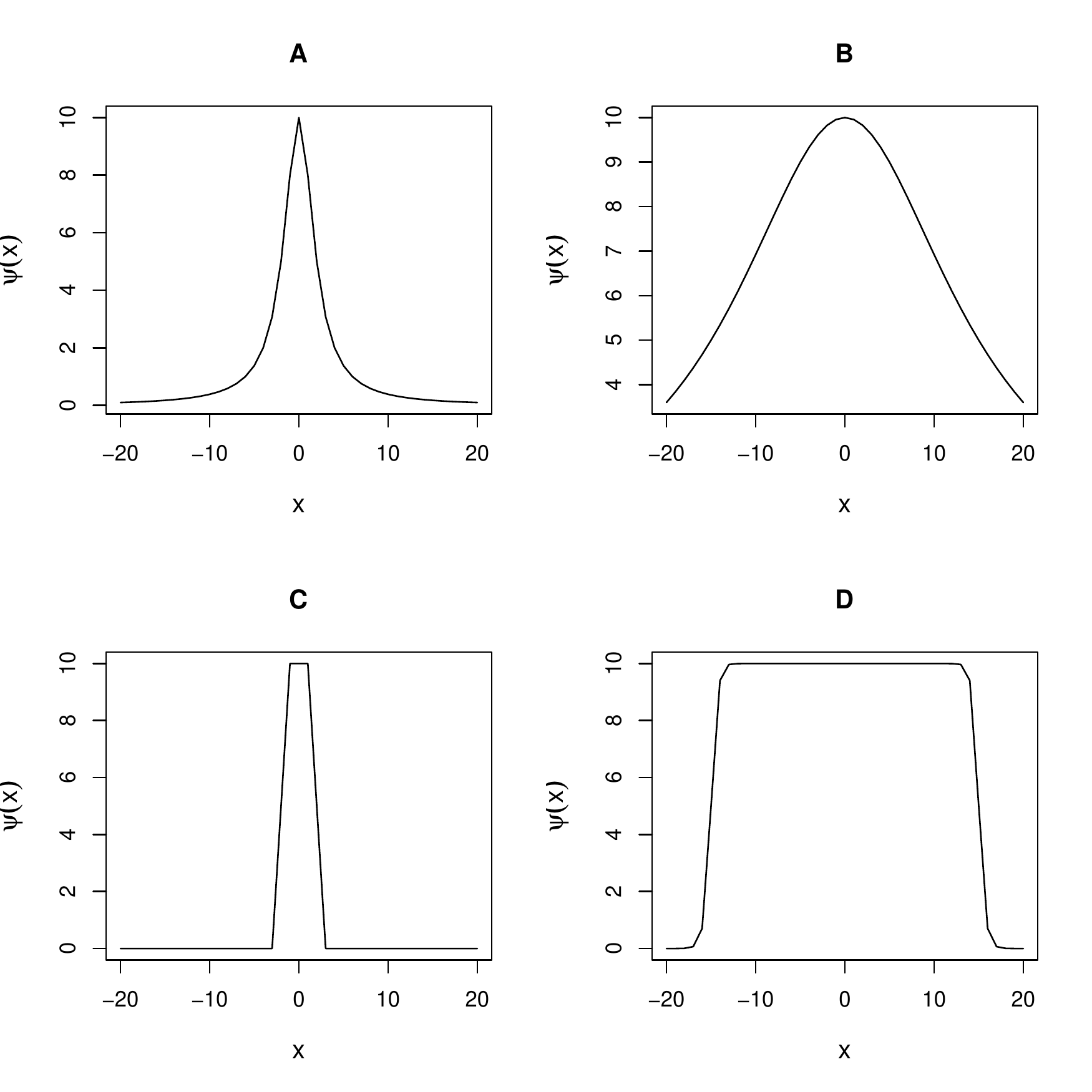,width=3.0truein}
}
\end{center}
\caption{ Shaping the tolerance function $\psi$. A: amplitude = 10, cutoff = 2, support = 0, order= 1. B: amplitude = 10, cutoff = 15, support = 0, order= 1. C: amplitude = 10, cutoff = 2, support = 0, order= 20. D: amplitude = 10, cutoff = 15, support = 0, order= 20.}
\label{fig3}
\end{figure}

\begin{figure}[t]
\begin{center}
\centerline{ 
 \psfig{file=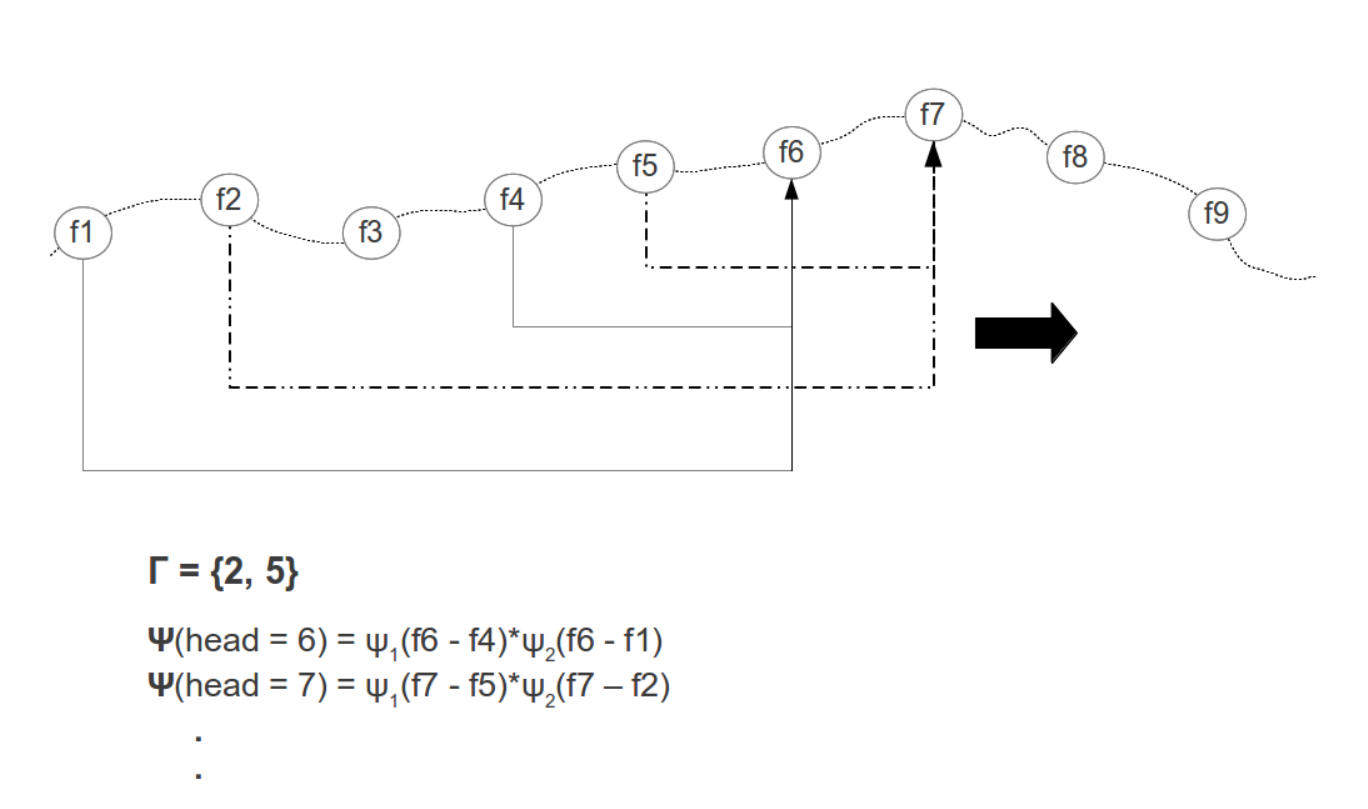,width=4.5truein}
}
\end{center}
\caption{The above figure illustrates the process by which a TPD scans the occurrences of a desired temporal pattern in a discrete signal. For the above TPD $\zeta$ = 2; hence the TPD consists of 3 pointers representing a trident. The vertical arrow (on the $3^{rd}$ arm of the trident) represents the current value of $x_{head}$. The horizontal arrow as well as the dotted-trident suggests that the TPD will be sequentially moved across the signal.}
\label{fig4}
\end{figure}

The temporal pattern detection task involves detecting predefined temporal structures in a time-series. Roy et al. \citep{Roy2013} introduced a design rule for a spike neural network based TPD which consisted of a serial chain of sequence detectors, each designed to detect the occurrences of a predefined inter-spike interval (ISI) pattern in a serial spike train, within a fixed tolerance limit (a box-car function). The design specifications were many fewer than the number of network parameters for the TPD. This provided an opportunity for the evolutionary algorithm to learn the design specifications, instead of having to tune myriad network parameters. This idea was successfully tested on the alcoholic classification task where characteristic temporal patterns in the alcoholic VERPs were found by first converting the time-series to a spike train \citep{Roy2013} and then evolving the TPDs to detect hidden ISI patterns. We tried evolving the above TPD for a larger alcoholic dataset, however, the results were not satisfactory. We hypothesize that due to the rigid tolerance window associated with the above TPDs, partial-credit cannot be assigned to the temporal patterns that do not fall within the desired specification, but come close to it. This may result in a fitness landscape that may not be favorable for the evolutionary process. Here, we will introduce a new temporal pattern detector where the tolerance window for the desired temporal pattern will be represented by a $\zeta$-dimensional continuous function:
\begin{equation}
 \Psi(x_{head}) = \prod_{\Gamma_{j} \in \Gamma} \psi_{j}(f(x_{head})-f({x_{head}-\Gamma_{j}})) 
 \end{equation}
 This enables us to rate the temporal patterns on a continuous scale, unlike the TPD introduced by Roy et al. \citep{Roy2013} where a boolean rating system was used. The function $\psi_{j}$ is formally explained below:
\begin{itemize}
\item[]{$f(x)$ represents a finite length ($\left|{f(x)}\right|$) discrete time-series, $\lbrace$ $x$ $\in$ $\mathbb{Z}$ $\vert$  $x$  $>$ 0 $\rbrace$, and $\lbrace$ $f(x)$ $\in$ $\mathbb{R}$ $\rbrace$.\\}
\item[]{$\Gamma$ represents a set of distinct pointers on the time-series where $j$ indexes an element, $\Gamma_{j}$, of the set. Also, $\left| \Gamma \right|$ $=$ $\zeta$, $\lbrace$ $\Gamma_{j}$  $\in$ $\mathbb{Z}$ $\rbrace$, and $\lbrace$ $\zeta$  $\in$ $\mathbb{Z}$ $\vert$  $\zeta$  $>$ 0 $\rbrace$.\\}
\item[]{$x_{head}$ represents the lead pointer to a position on $f(x)$, such that $\lbrace$ $x_{head}$  $\in$ $\mathbb{Z}$ $\vert$ $max(\Gamma_{j})_{\Gamma_{j} \in \Gamma}$ $<$ $x_{head}$  $\le$ $\left|{f(x)}\right|$ $\rbrace$.\\}
\end{itemize}
\begin{equation}
\psi_{j}(f(x_{head})-f(x_{head}-\Gamma_{j})) = \frac{amplitude_{j}}{(1 + (\frac{(f(x_{head})-f(x_{head}-\Gamma_{j})) - support_{j}}{cutoff_{j}})^{2*order_{j}})}
\end{equation}
where, $\lbrace$ $amplitude_{j}$ $\in$ $\mathbb{R}$ $\rbrace$, $\lbrace$ $support_{j}$  $\in$ $\mathbb{R}$ $\rbrace$, $\lbrace$ $cutoff_{j}$  $\in$ $\mathbb{R}$ $\vert$  $cutoff_{j}$  $>$ 0  $\rbrace$, and 
$\lbrace$ $order_{j}$  $\in$ $\mathbb{Z}$ $\vert$   $order_{j}$  $\ge$ 1 $\rbrace$.
 \\\\
\textbf{Summarizing the function $\psi_{j}$\\\\}
The function $\psi_{j}$ is used as a tolerance function for the amplitude difference between the points, $x_{head}$ and $\Gamma_{j}$, where, $support_{j}$ represents the desired amplitude difference (a part of the desired pattern). It reaches its maximum value, $amplitude_{j}$, when $(f(x_{head})-f(x_{head}-\Gamma_{j})) = support_{j}$. The parameters $cutoff_{j}$ and $order_{j}$ are used for controlling the width and the manner in which $\psi_{j}$ decays (see Figure~\ref{fig3}), respectively. The shape of the function $\psi_{j}$ influences the partial credit assignment for the temporal patterns that deviate from the desired temporal structure. The product, $\prod_{\Gamma_{j} \in \Gamma} \psi_{j}(f(x_{head})-f({x_{head}-\Gamma_{j}}))$ is a mechanism by which we can evaluate to what degree the individual amplitude differences between the points, $x_{head}$ and $\Gamma_{j}$, have deviated from the desired set of amplitude differences (the desired temporal pattern). The occurrences of the desired temporal pattern in a discrete signal $f(x)$ can be evaluated by moving the pointer $x_{head}$ sequentially over the discrete-signal (see Figure~\ref{fig4}). This can be formally stated as follows: 
\begin{equation}
 \Phi = \sum_{ x_{head} = 1 + max(\Gamma_{j})_{\Gamma_{j} \in \Gamma}}^{\left|{f(x)}\right|}   \Psi(x_{head})
\end{equation}



The element(s) of the set $\Gamma$, and the parameters $amplitude_{j}$, $cutoff_{j}$, $order_{j}$, and $support_{j}$, will all be set by the evolutionary process. For the purposes of this paper we will keep the function $\Psi(x_{head})$ symmetric, i.e. for all the possible values of $j$, the parameters $amplitude_{j}$, $cutoff_{j}$, and $order_{j}$, will be set to the evolved values, $amplitude$, $cutoff$, and $order$, respectively. The $support_{j}$ value(s) will be set using an indirect encoding scheme, explained later in the paper.



\begin{figure}[t]
\begin{center}
\centerline{ 
 \psfig{file=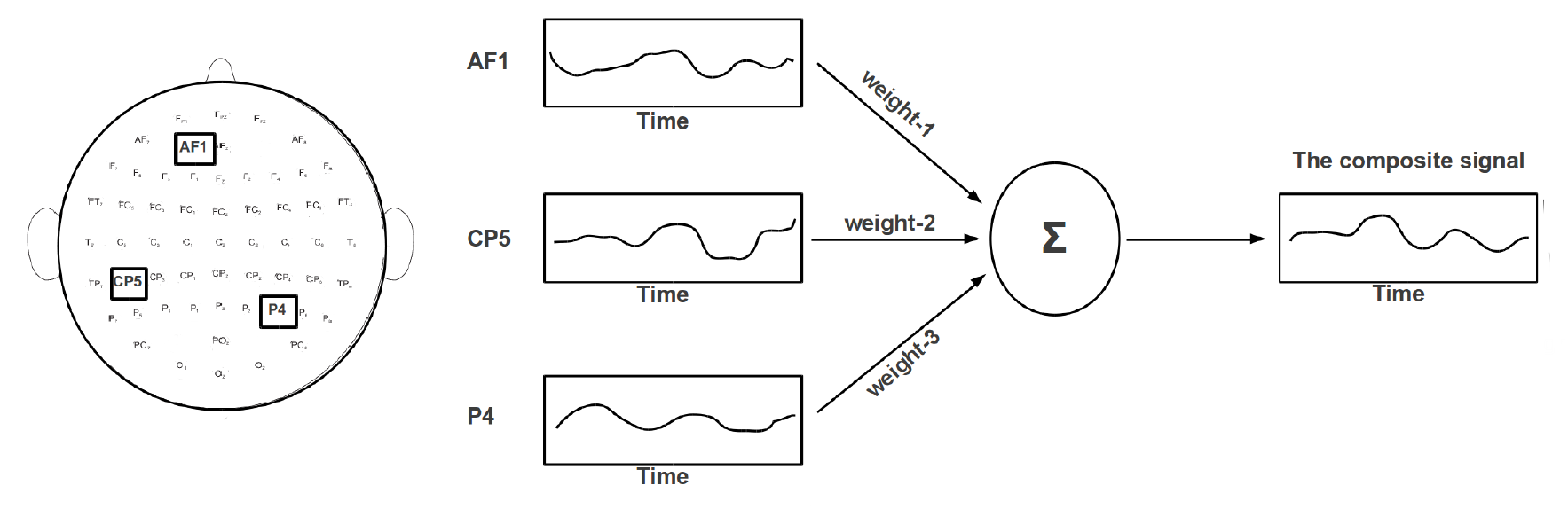,width=4.5truein}
}
\end{center}
\caption{The above figure illustrates the steps involved in the creation of a composite signal. The leftmost figure represents the EEG electrodes. In order to create the composite signal, electrodes AF1, CP5, and P4 are selected. The second column shows the signal arriving from the chosen electrodes. Finally, the weighted sum of these signals, the composite signal, is shown in the last column.}
\label{fig5}
\end{figure}

\subsection{The chromosome encoding}
The alcoholic classification problem consists of 2 sub-tasks: the spatial task, and the temporal task. The spatial task involves choosing an appropriate subset of EEG leads along with the lead-weights using which a composite signal can be created (see Figure~\ref{fig5}). The objective of the temporal task is to design a TPD that can find a characteristic temporal pattern in the alcoholic composite signals; these patterns should occur more frequently in the alcoholic VERPs, and many fewer times in the control VERPs. Since, there are only 62 EEG leads, a sub-chromosome is assigned to directly encode (the genotype is the phenotype) the leads along with the weights; an EEG lead is considered a feature and the corresponding weight is its numeric (real number) attribute. Contrastingly, the space of possible temporal patterns is overwhelmingly large. As a consequence, the TPD design specifications are indirectly encoded into 3 sub-chromosomes; the sub-chromosomes will encode a teacher-id (a pointer to an alcoholic training signal) and a specific set of pointers to the positions on the teacher composite signal (see Table~\ref{table2}). The signal amplitudes at these positions will define the temporal pattern the TPD will be designed to detect; this guarantees that the temporal pattern of interest exists in at-least one alcoholic composite signal and may qualify as a candidate solution for the classification task. This approach involves an implicit assumption that the alcoholic signals are homogeneous. In order to accommodate the possibility of class heterogeneity, we allow a chromosome to encode at most 2 teacher signals using which 2 TPDs can be created. This decision was based on the outcome of a preliminary investigation, where the objective was to establish a sense of the minimum number of TPDs required \footnote{A classifier composed of too many TPDs may over-fit the training set.} in order for the classifier to distinguish the alcoholic cases with an acceptable precision and accuracy. The information encoded by the sub-chromosomes are summarized in Table~\ref{table2}:

\begin{table}[h]
\begin{center}

\begin{tabular}{|c|p{1.8 cm}|p{2 cm}|p{5 cm}|c|}
 \hline
\tiny{\bf Sub-chromosome}  & \tiny{\bf Feature-id } &  \multicolumn{1}{c|}{\tiny{\bf Attribute(s)}} &  \multicolumn{1}{c|}{\tiny{\bf Comments}}&  \tiny{\bf subset length } \\ 
  & \tiny{\bf (Max/Min)} & \tiny{\bf (Max/Min/type)} &  \multicolumn{1}{c|}{}&  \tiny{\bf Max/Min} \\ \hline\hline
\tiny 1  &  \tiny Sensor-id (62/1)  & \raggedright \tiny Sensor-weight\\(4.0/-4.0/Real) & & \tiny 5/1 \\ \hline

\tiny 2  &  \tiny Teacher-id (47/1)  &  \tiny  & \raggedright \tiny{There are 47 alcoholic training signals each index by an unique id.} & \tiny 2/1 \\ \hline

\tiny 3  &  \tiny Reference-pointer (250/97)  & \raggedright \tiny Skip-length\\(12/1/Int) & \raggedright \tiny{1. The reference pointer (RP) marks a position on the composite teacher signal w.r.t which other pointers will be defined. The signal amplitude at these positions will then define the temporal pattern of interest (TPI).\\ \vspace{5pt}2. The teacher signal amplitude at the positions $(RP - \varphi_{j}\ast$ $skip$-$length)$ are candidates for defining the TPI, where $\lbrace$ $\varphi_{j}$  $\in$ $\mathbb{Z}$ $\vert$ 1  $\le$ $\varphi_{j}$ $\le$ 8 $\rbrace$.} & \tiny 2/1 \\ \hline

\tiny 4  &  \tiny Qualification-id (255/1) & \raggedright \tiny Cutoff\\(20.0/0.1/Real)\\ \vspace{5pt} Order\\(15/1/Int)\\ \vspace{5pt} Amplitude\\(1.0/0.0/Real) & \raggedright \tiny{1. Qualification-id (QI) is first converted to a 8-bit binary form. If the value at bit position $j$ is 1, then $\varphi_{j} = j$ else $\varphi_{j}$ doesn't qualify for defining the TPI.\\ \vspace{5pt}2. The Cutoff, Order, and the Amplitude values are same for all $\psi_{j}$ (see TPD equations).} & \tiny 2/1 \\ \hline

 \hline 
\end{tabular}
\caption{\label{table2} The above table illustrates the chromosome encoding scheme for the alcoholic classification task. The sub-chromosome-1, sub-chromosome-2, and sub-chromosome-3, must be of same length. }
\end{center}
\end{table}


\subsection{The objective function/selection procedure}

The aim of the evolutionary process is to find a TPD whose output ($\Phi$) will be much larger for the alcoholic composite signals than the control composite signals. In order to assign a penalty-value to a chromosome, the output of the TPD (that it encodes) for all the training cases (47 alcoholic + 31 control cases) are stored as two distributions: $\Phi_{alcoholic}$ and $\Phi_{control}$. These discrete-distributions are then used for calculating the area under the receiver operator characteristic curve (AUC). The penalty is then evaluated as follows:

\begin{equation}
penalty = 1 - AUC
\end{equation}

Thus, penalty = 0 would suggest that the output of the TPD, $\Phi$, for the alcoholic cases are always greater than the control cases, implying there exists no overlap between the $\Phi_{alcoholic}$ and the $\Phi_{control}$ distributions.\par We have implemented an elitist selection procedure introduced by Eshelman \citep{Eshelman1991}, where the parents and the offspring compete for a position in the population. The parents and the offspring are first sorted by penalty (in ascending order), and only the top $\rho$ chromosomes are selected \footnote{The fittest chromosomes are selected}, where $\rho$ represents the population size. Since, the subset sizes for the current application were modest, no selection pressure to evolve smaller subsets was deemed necessary.

\begin{figure}[t]
\begin{center}
\centerline{ 
 \psfig{file=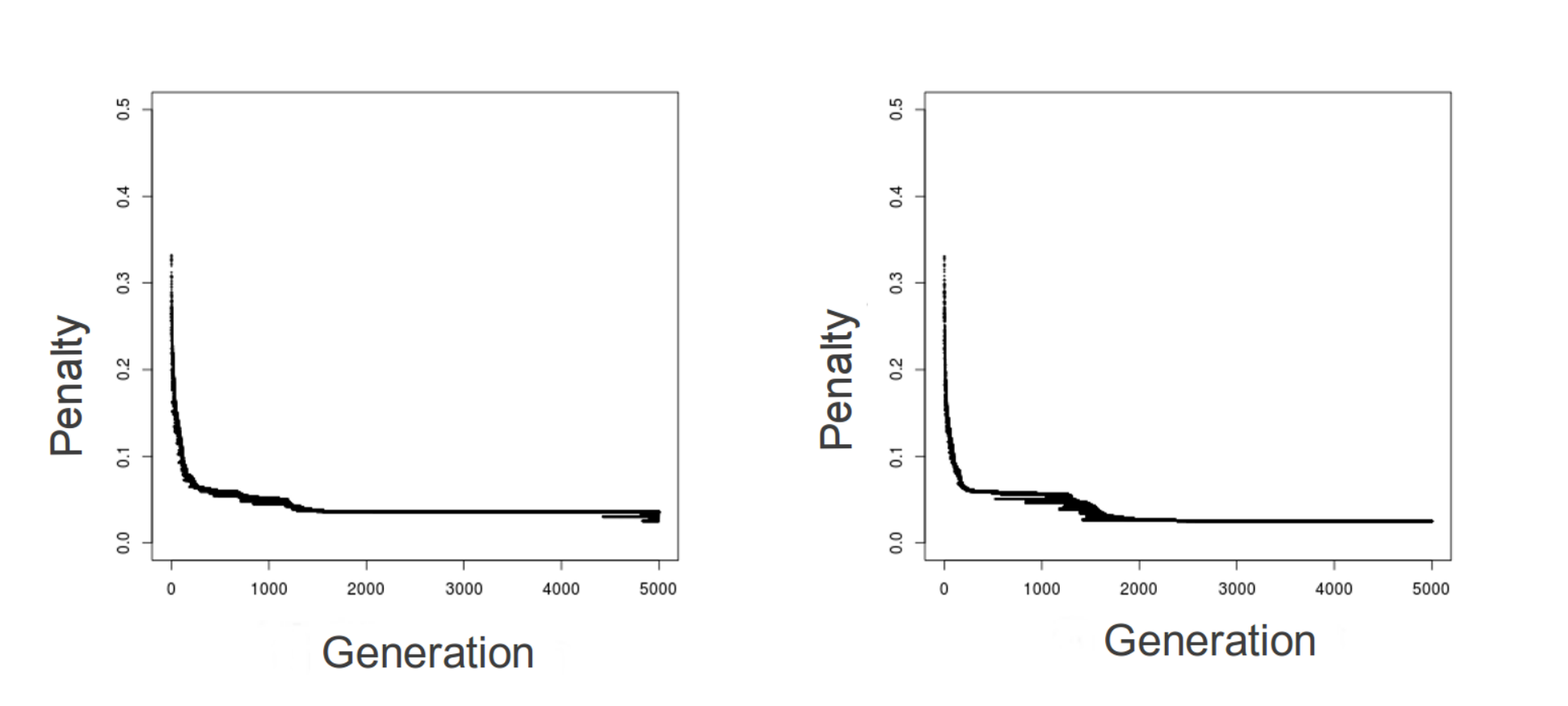,width=4.5truein}
}
\end{center}
\caption{The above plot illustrates how the population penalty changes over generations when MMX-BLX$^{explore}$ (Left) and MMX-BLX$^{exploit}$  crossover operators (right) are implemented.}
\label{fig6}
\end{figure}


\begin{figure}[t]
\begin{center}
\centerline{ 
 \psfig{file=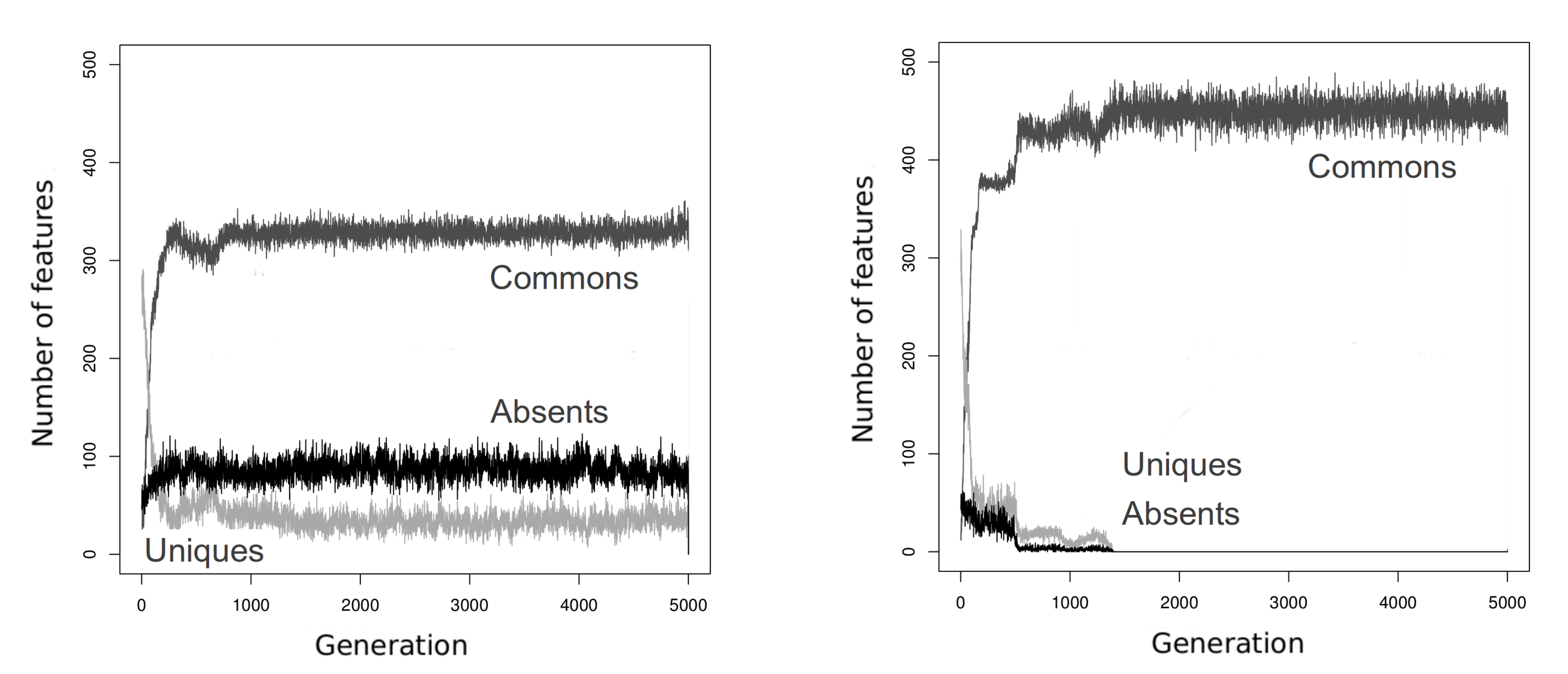,width=4.5truein}
}
\end{center}
\caption{The above plot illustrates how the number of features that are selected from the bag of commons, the bag of uniques, and the bag of absents change  over generations when MMX-BLX$^{explore}$  (left) and MMX-BLX$^{exploit}$  crossover operators (right) are implemented. A mutation event involves selecting a feature from the bag of absents.}
\label{fig7}
\end{figure}


\begin{figure}[t]
\begin{center}
\centerline{ 
 \psfig{file=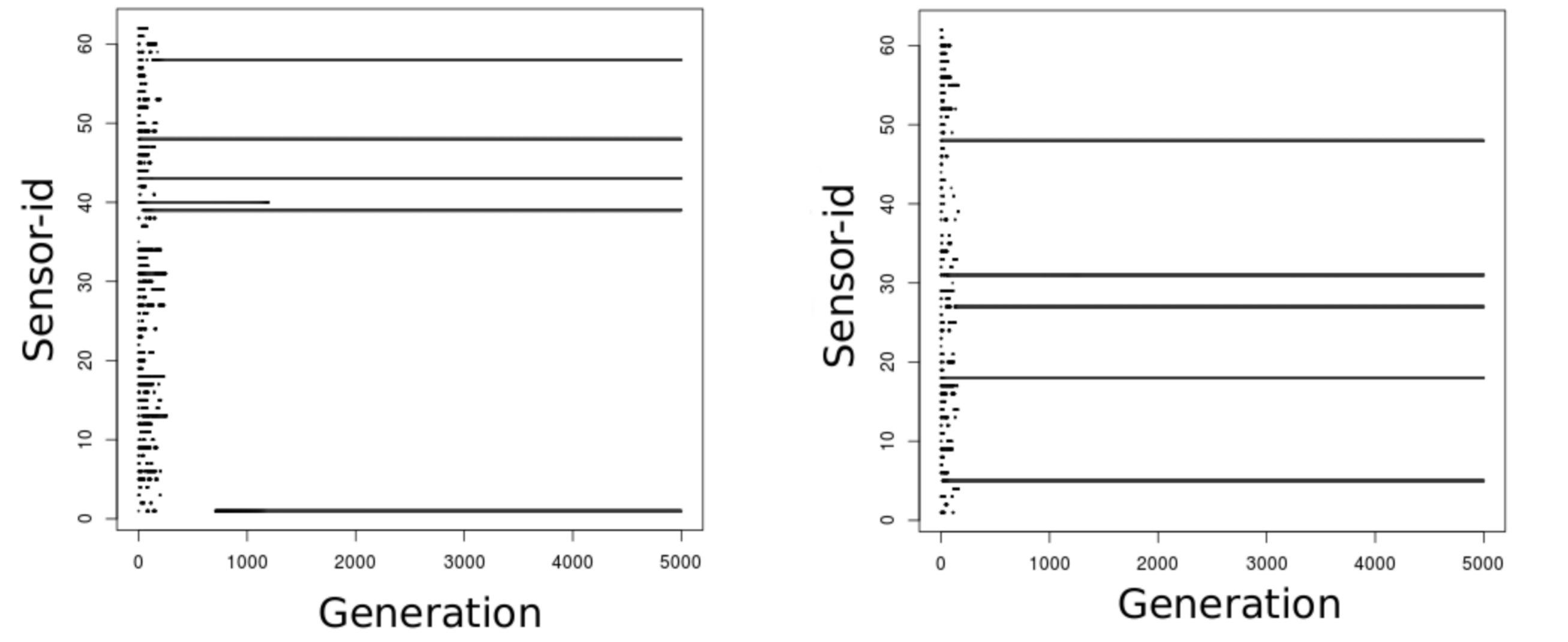,width=4.5truein}
}
\end{center}
\caption{Emergence of sensor-subsets in two evolutionary runs: MMX-BLX$^{explore}$  (left) and MMX-BLX$^{exploit}$ (right). In both cases a sensor subset of size 5 was evolved.}
\label{fig8}
\end{figure}

\section{Results}

In order to characterize the behavior of  MMX-BLX$^{explore}$ and MMX-BLX$^{exploit}$, as well as to test the repeatability of the learning paradigm, we conducted 3 independent experiments using each crossover operator. The population size was set to 50 and each experiment was run for 5000 generations. The crossover parameters $\alpha$, $\beta$, $\delta$, and $\gamma$, were set to 1, 1.4, 0.85, and 0.75, respectively. In Figure~\ref{fig6}, Figure~\ref{fig7}, and Figure~\ref{fig8}, we have illustrated the behavior of these crossover operators for only one experiment; the behavior was similar across all 3 experiments. The performance of the evolved classifier for all 3 experiments have been summarized in Table~\ref{table3}. For both crossover operators the population penalty had an initial rapid fall till generation-400 (see Figure~\ref{fig6}) during which the evolution made a decision on the spatial aspect of the problem; chromosomes encoding undesirable combinations of EEG sensors were eliminated (see Figure~\ref{fig8}). Even though the chosen subset of sensors varied between the experiments, they primarily represented the central, the right-parietal/occipital, and the right-frontal regions of the brain. Once the spatial task was solved, from generation-400 onwards the evolution focused on the temporal aspect of the problem. We did not predefine the order in which these tasks should be solved; it was an emergent behavior.\par Both the operators were able to produce a good solution by $2500^{th}$ generation. While the population penalty for MMX-BLX$^{exploit}$ remained unchanged after generation-2500, MMX-BLX$^{explore}$ was able to find new solutions between generation-4500 to generation-5000, suggesting that the current task may require more exploration. For MMX-BLX$^{exploit}$, after generation-1500 there were no mutation events; all features were chosen from the bag of commons (see Figure~\ref{fig7}). Also the mutation events gradually decreased to 0 by generation-1500. This corroborates our hypothesis that MMX-BLX$^{exploit}$ exhibits a CCM and is less exploratory in nature. Figure~\ref{fig7} also suggests that MMX-BLX$^{explore}$ has a quasi-constant mutation rate that helps maintain genetic diversity.\par In Table~\ref{table3} we have illustrated the performance of the best chromosome in generation-5000 on both training and the test cases. Interestingly, in spite of the fact that MMX-BLX$^{exploit}$ is less exploratory in nature, it performed slightly better than MMX-BLX$^{explore}$ on the training set. However, MMX-BLX$^{explore}$ was able to find more robust solutions; its performance on the test set was much better. Perhaps, MMX-BLX$^{exploit}$ was handicapped by not having a soft-restart mechanism \citep{Eshelman1991}. Such a mechanism requires an incest prevention scheme based on a similarity metric that we did not implement as the chromosome-structure was complex.

\begin{table}[h]
\begin{center}

\begin{tabular}{|c|p{3 cm}|p{3 cm}c|}
 \hline
\tiny{\bf Experiment No.}   &  \multicolumn{1}{c|}{\tiny{\bf  MMX-BLX$^{explore}$}} &  \multicolumn{1}{c}{\tiny{\bf MMX-BLX$^{exploit}$}}& \\ 
\tiny{\bf}   &  \multicolumn{1}{c|}{\tiny{\bf  Training/Test penalty}} &  \multicolumn{1}{c}{\tiny{\bf Training/Test penalty}}& \\ \hline\hline

\tiny 1  &\raggedright \tiny{0.0254/0.0417} & \raggedright \tiny{0.0254/0.1282} &  \\ \hline
\tiny 2  &\raggedright \tiny{0.0357/0.0703} & \raggedright \tiny{0.0295/0.09} &  \\ \hline
\tiny 3  &\raggedright \tiny{0.046/0.0906} & \raggedright \tiny{0.0336/0.1283} &  \\ \hline

 \hline 
\end{tabular}
\caption{\label{table3} The above table summarizes the performance of the evolved classifier. }
\end{center}
\end{table}

\section{Conclusion}

Both, MMX-BLX$^{explore}$  and MMX-BLX$^{exploit}$ provides a mechanism to simultaneously solve multiple FS problems where the features may have numeric attribute(s) and the subset size is not predefined. Since, MMX-BLX$^{exploit}$ prioritizes positive respect it is able to perform rigorous search in a region defined by the parental chromosomes. On the one hand this feature allows it to perform better on the training set, but on the other hand it seems to yield less general solutions. A soft-restart mechanism may allow MMX-BLX$^{exploit}$ to perform more exploratory search. \par MMX-BLX$^{explore}$ seem to be able to find more robust solutions by virtue of quasi-constant mutation process even-though the fitness function used did not explicitly evaluate generalization. Mutation allows more exploration by sacrificing respect; this trade-off may be problem specific and requires additional investigation.\par The conventional techniques used for distinguishing the alcoholic VERPs from the controls primarily consist of 2 steps, developing a set of feature vectors and training a classifier using these feature vectors \citep{pala2002, pala2005, pala2006, pala2007, okming2005, Kousarrizi2009, Shri2012, Shahina2008}. Most authors have used the information in the gamma band (30-50 Hz) to develop feature vectors. Using, the evolutionary learning paradigm along with the TPD technology we were able to solve this  problem in 1 step; we did not make any assumptions regarding the data. The TPD technology introduced here is an extension of the pattern detector developed by Roy et al. \citep{Roy2013} and  provides a mechanism to assign partial credits to temporal patterns that vary from the desired specification. We believe this makes the search landscape more favorable to an effective evolutionary search. Finally, by allowing the evolution to interact with the environment (the alcoholic teacher signal), the search for a temporal pattern that can characterize the alcoholic VERP is made more manageable.

\section{Acknowledgements}
The data for this research was made available on the web by Henri Begleiter, Neurodynamics Laboratory, State University of New York Health Center at Brooklyn.

\small

\bibliographystyle{apalike}

\end{document}